\documentclass[]{ceurart}

\usepackage{algorithm}
\usepackage{algorithmic}
\usepackage[framemethod=TikZ]{mdframed}
\usepackage{subfigure}
\usepackage{enumitem}

\begin{document}

\copyrightyear{2025}
\copyrightclause{Copyright for this paper by its authors.
  Use permitted under Creative Commons License Attribution 4.0
  International (CC BY 4.0).}

\conference{De-Factify 4.0: Workshop on Multimodal Fact Checking and Hate Speech Detection, co-located with AAAI 2025}

\title{Don't Fight Hallucinations, Use Them: Estimating Image Realism using NLI over Atomic Facts}

\author[1]{Elisei Rykov}[
email=Elisei.Rykov@skol.tech
]
\cormark[1]

\author[1,2]{Kseniia Petrushina}[
email=Kseniia.Petrushina@skol.tech
]
\cormark[1]

\author[1,3]{Kseniia Titova}[
email=Kseniia.Titova@skol.tech
]

\author[1,4]{Alexander Panchenko}[
email=a.panchenko@skol.tech
]

\author[4,2]{Vasily Konovalov}[
email=vasily.konovalov@phystech.edu
]
 
\address[1]{Skoltech}
\address[2]{Moscow Institute of Physics and Technology}
\address[3]{MTS AI}
\address[4]{AIRI}

\cortext[1]{These authors contributed equally.}

\begin{abstract}
  Quantifying the realism of images remains a challenging problem in the field of artificial intelligence. For example, an image of Albert Einstein holding a smartphone  violates common-sense because modern smartphone were invented after Einstein's death. We introduce a novel method for assessing image realism using Large Vision-Language Models (LVLMs) and Natural Language Inference (NLI). Our approach is based on the premise that LVLMs may generate hallucinations when confronted with images that defy common sense. Using LVLM to extract atomic facts from these images, we obtain a mix of accurate facts and erroneous hallucinations. We proceed by calculating pairwise entailment scores among these facts, subsequently aggregating these values to yield a singular reality score. This process serves to identify contradictions between genuine facts and hallucinatory elements, signaling the presence of images that violate common sense. Our approach has achieved a new state-of-the-art performance in zero-shot mode on the WHOOPS! dataset.
\end{abstract}


\maketitle

\section{Introduction}

When presented with an unusual image, human perception quickly detects discordant elements. For example, an image of Einstein holding a smartphone can appear ordinary in their components yet strikingly abnormal in their arrangement. While humans intuitively spot the non-sense of the image, the cognitive process behind this is intricate. Linking visual elements to common-sense extends beyond simple object recognition~\cite{zellers2019recognition}.

\begin{figure*}[t]
  \centering
    \begin{minipage}{0.49\textwidth}
        \centering
        \small
        \includegraphics[width=0.6\textwidth]{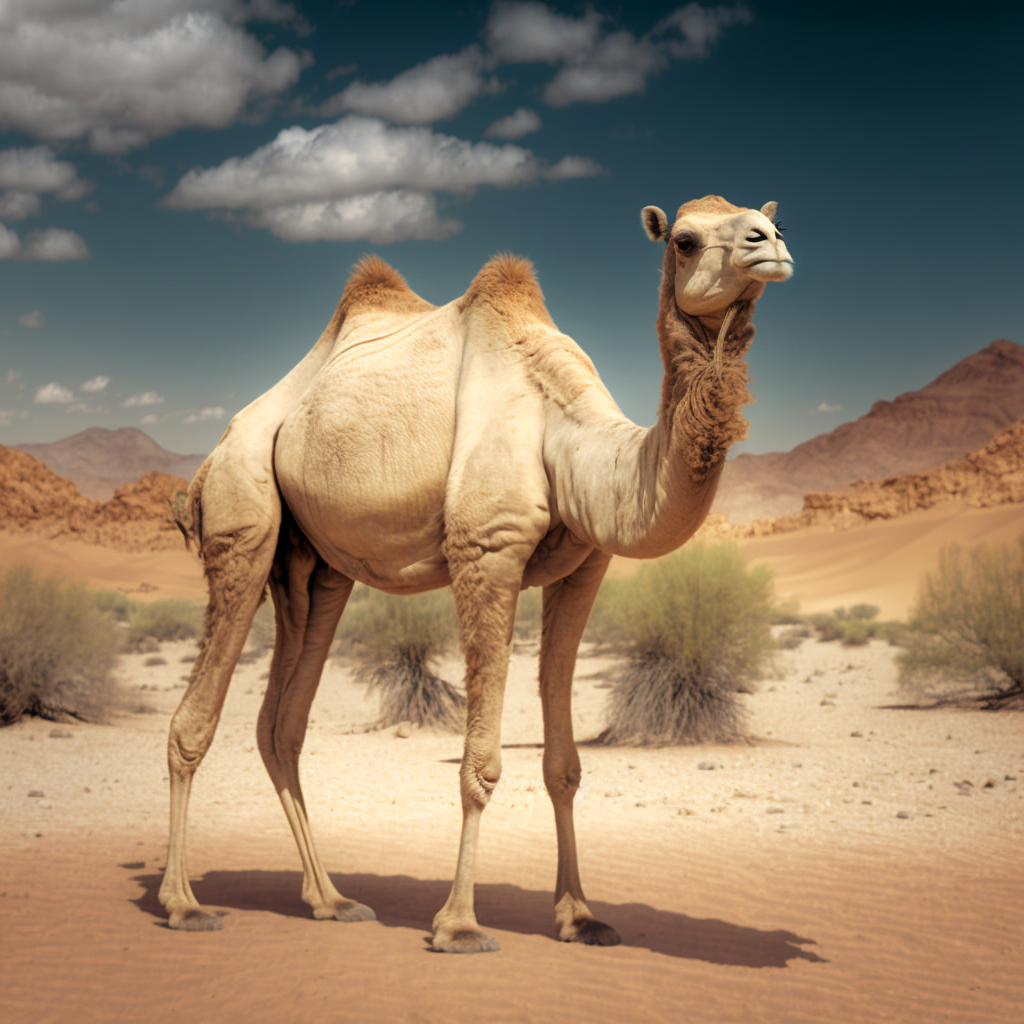}\\
        \begin{mdframed}
            \textit{This is a camel}\\
            \textit{The camel is standing on sand}\\
            \textit{This image features a camel standing on a sandy desert plain}
        \end{mdframed}
    \end{minipage}\hfill
    \begin{minipage}{0.49\textwidth}
        \centering
        \small
        \includegraphics[width=0.6\textwidth]{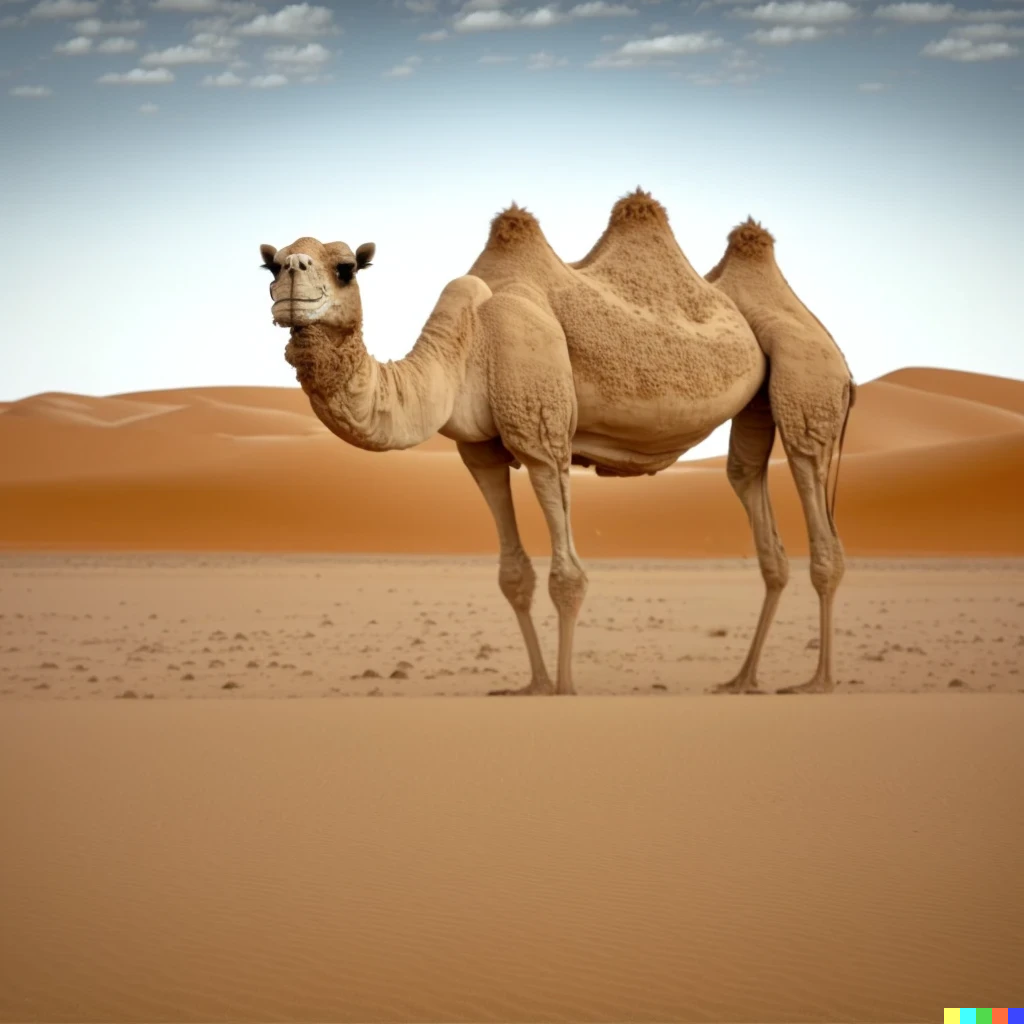}\\
        \begin{mdframed}
            \textit{The camel is in the desert}\\
            \textit{The \textcolor{red}{camels} are standing in the desert}\\
            \textit{This is a digitally manipulated image of a camel with \textcolor{red}{two heads}}
        \end{mdframed}
    \end{minipage}
    \caption{A pair of images from the WHOOPS! dataset with corresponding generated atomic facts. The normal image is on the left, and the unusual image is on the right. All the facts associated with the normal image are consistent and accurately describe the image. However, in the case of the weird image, LVLM hallucinates and generates untruthful facts.}
    \label{mainpicture}
\end{figure*}

In this work, we propose a visual common sense zero-shot classifier that utilizes observation that LVLMs may generate hallucinations when confronted with images defying common sense~\cite{phd}. Using LVLMs to extract atomic facts from these images, we obtain a mix of accurate facts and erroneous hallucinations. Then we calculate pairwise entailment scores among these facts, subsequently aggregating these values to produce a singular reality score (Figure~\ref{mainpicture}). This process serves to identify contradictions between genuine facts and hallucinatory elements, signaling the presence of images that violate common sense.

Our findings suggest that rather than relying on intricate models, we can effectively exploit the imperfections of LVLMs – their tendency to generate hallucinations – in conjunction with encoder-based NLI models, which were found to be effective in detecting hallucinations~\cite{maksimov-etal-2024-deeppavlov}, to identify images that defy common sense. Interestingly, our approach surpasses all zero-shot open solutions and demonstrates performance comparable to fine-tuned baseline methods (BLIP-2 FlanT5-XXL)~\cite{whoops}.

Our contributions can be summarized as two-fold:
\begin{itemize}[noitemsep]
\item First, we confirm the previous observation that LVLMs may generate hallucinations when confronted with images that contradict common sense.
\item Second, leveraging this observation, we propose a simple yet effective NLI-based common sense visual classifier that achieves state-of-the-art performance among open-source models in zero-shot mode.\footnote{\url{https://github.com/s-nlp/dont-fight-hallucinations}}
\end{itemize}

\section{Related Work}

Recently, the realm of commonsense reasoning has attracted substantial interest, spanning across disciplines within Natural Language Processing (NLP) and computer vision (CV), with numerous tasks being introduced. ~\citet{whoops} introduced the WHOOPS! 
benchmark, comprising images specifically designed to challenge common sense. In an effort to detect unconventional images, they employed BLIP-2~\cite{li2023blip2} Flan-T5 at multiple scales. Although the fine-tuned model managed to outperform a random baseline, it still falls significantly short of human performance.

\citet{realistic_image} provided a theoretical basis for why quantifying realism is challenging. He suggests considering not just the visual fidelity of images, but also the contextual and common-sense coherence. An image might appear visually convincing but still fail to align with our understanding of how the world works, thereby diminishing its realism.

~\citet{phd} introduced a novel PhD (ChatGPT \underline{P}rompted visual \underline{h}allucination evaluation \underline{D}ataset) benchmark, which includes a subset of counter-common-sense (CCS) images. Each image in the set is accompanied by two yes/no questions pertaining specifically to the image.

\citet{faithscore} proposed a benchmark and the FAITHSCORE method that measures faithfulness of the generated free-form answers from LVLMs. The FAITHSCORE first identifies sub-sentences containing descriptive statements that need to be verified, then extracts a comprehensive list of atomic facts from these sub-sentences, and finally conducts consistency verification between fine-grained atomic facts and the input image. Our approach resembles the preceding method in that we also derive atomic facts from the image by using LVLMs.

\section{Problem Statement}

The task is to obtain a quantitative score of the realism of a particular image. Given two sets with usual and weird images that contradict common sense, we need to find a method to separate real images from weird ones. We propose \textit{reality-check} function $\text{RealityCheck}( \cdot )$, which returns the image reality score. For real image $I_r$ and weird image $I_w$ we get

\begin{equation}
    \text{RealityCheck}(I_r) > \text{RealityCheck}(I_w)
\end{equation}

\section{Proposed Method}

\begin{figure*}[hbt!]
\centering
\includegraphics[width=0.95\textwidth]{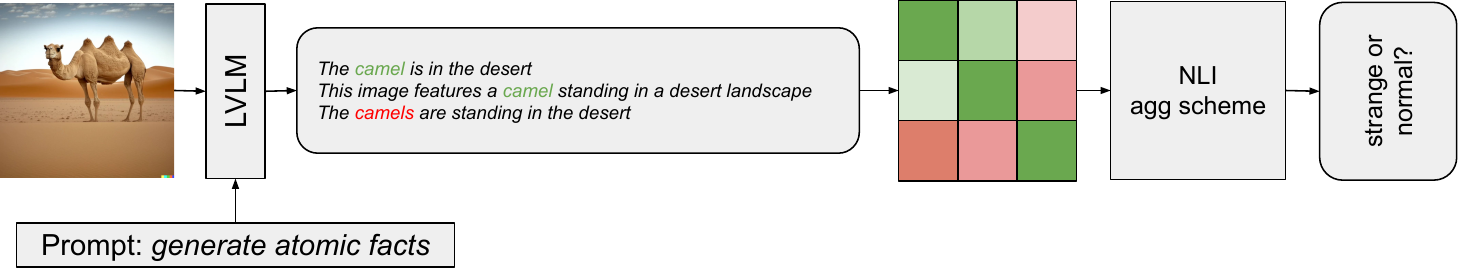}
\caption{Weird images detection pipline. First, we generate five atomic facts that describe the images with LVLM (\textsf{\small{llava-v1.6-mistral-7b-hf}}). Then, we proceed with the matrix of pairwise NLI scores, where each NLI score is a weighted combination of entailment, neutral, and contradiction scores. The last step is aggregating NLI scores. Then, based on the aggregated score, we decide whether the image is strange or not.}
\label{pic:pipeline}
\end{figure*}

Our RealityCheck method is based on three steps: (i) we prompt the LVLM to generate multiple atomic facts describing the image; (ii) for each pair of atomic facts, we get the NLI scores; (iii) finally, we aggregate the NLI scores into RealityCheck score. The pipeline of our approach is depicted in Figure~\ref{pic:pipeline}.

\subsection{Fact Generation}

The first step of our approach is to generate a list of atomic facts about the input image $F = \{ \text{LVLM}(I, P) \}_{i=1}^N,$
where $I$ -- an input image, $P$ -- a textual prompt to generate simple atomic facts about the image. We generate $N$ different facts using the Diverse Beam Search~\cite{DBLP:journals/corr/VijayakumarCSSL16}. We employ \textsf{\small{llava-v1.6-mistral-7b-hf}}\footnote{\url{https://hf.co/llava-hf/llava-v1.6-mistral-7b-hf}} model for diverse fact generation\footnote{Generation was performed on a single V100 32GB GPU.} with the following prompt: ``\textit{Provide a short, one-sentence descriptive fact about this image.}". Specifics on the generation parameters are presented in the Appendix~\ref{app:fact}.

\subsection{Pairwise NLI}

For each pair of facts $(f_i, f_j)\in F \times F$, we get entailment, contradiction, and neutrality scores from the NLI model:

\begin{equation}
s_{\text{ent}},\;s_{\text{con}},\;s_{\text{neu}} = \text{NLI}(f_i, f_j)
\end{equation}

Then we aggregate the different scores into a single $\text{s}_\text{nli}$ using weighted aggregation:

\begin{equation}
s_\text{nli}(f_i, f_j) = w_{ent} \cdot s_{ent} + w_{con}\cdot  s_{con} + w_{neu} \cdot s_{neu}
\end{equation}

\subsection{NLI Score Aggregation}
After aggregation and weighting, we calculate a sum of weighted scores for each pair of facts $s_\text{nli}(f_i, f_j)$ and $s_\text{nli}(f_j, f_i)$. We hypothesize that such a summation strategy will amplify negative contradictions and vice versa. In addition, we propose three strategies for aggregating the NLI score.

\begin{equation}
S_\text{nli} = \big\{ s_\text{nli}(f_i, f_j) + s_\text{nli}(f_j, f_i) \mid  i, j \in \{1, \dots, N\}, \, i \neq j \big\}
\end{equation}

\noindent\textbf{Minimum (\textsf{\small min})}
For a given list of scores, we simply select the lowest value as the metric. We assume that the lowest value could represent the contradictory of the whole set of facts.
\begin{equation}
\text{RealityCheck}_\text{min} = \min(S_\text{nli})
\end{equation}

\noindent \textbf{Absolute maximum (\textsf{\small absmax})}
We transform all values from the list of scores to their absolute values, then select the index of the largest absolute value and return the value from the original list to preserve the sign of the original value. So, if some set of facts has a relatively strong contradiction, we choose it as a weird image and vice versa:
\begin{equation}
\text{RealityCheck}_\text{absmax} = S_\text{nli}\left[ \arg\max \left( \left| S_\text{nli} \right| \right) \right]
\end{equation}

\noindent \textbf{Clustering (\textsf{\small clust})}
We run the $k$-means clustering algorithm on the set of individual scores to split them into 2 clusters and then select the centroid with the lowest value as the metric. The algorithm can be found in the appendix~\ref{app:clust}. The choice of 2 clusters corresponds to the binary classification task. The idea is similar to the \textsf{\small min} method, but instead of the lowest value over all, we select an average of the values from the lowest cluster. We expect that contradictory facts from the weird images will have lower cluster centers than a related one.

Since NLI based on a cross-encoder architecture demonstrates its robustness and superior performance~\cite{sbert}, we adopt three cross encoders of different sizes: small
, base
, and large\footnote{\url{https://hf.co/cross-encoder/}}. All of these models have been tuned in a similar setup on the SNLI~\cite{snli} and MultiNLI~\cite{mnli} datasets.

\section{Experimental Setup}
In this section, we describe the details of our experimental setup: the baselines, the WHOOPS! dataset used for evaluation, and the main metric.

\subsection{Baseline}
First, we compare our proposed method with the WHOOPS baselines! benchmark. More specifically, the fine-tuned \textsf{\small{BLIP2 FlanT5-XL}}
and \textsf{\small{BLIP2 FlanT5-XXL}}\footnote{\url{https://hf.co/Salesforce/}} in cross-validation format. We also tested the \textsf{\small{BLIP2 FlanT5-XXL}} model in zero-shot format.

For the instruction-following baseline, we leverage LVLM with the prompt, which was found to be effective in detecting weird images~\cite{DBLP:conf/cvpr/LiuLLL24}: \textit{``Is this unusual? Please explain briefly with a short sentence.''} For evaluation, we selected \textsf{\small{LLaVA 1.6 Mistral 7B}} and \textsf{\small{LLaVA 1.6 Vicuna 13B}} with \textsf{\small{InstructBLIP}} of two sizes \textsf{\small{7B}} and \textsf{\small{13B}}.

\subsection{Dataset}
To evaluate our methods, we used WHOOPS! benchmark, focusing on a subset comprising 102 pairs of weird and normal images. Performance was measured by binary accuracy within this paired dataset, where a random guess would yield 50\% accuracy. To assess human performance, three annotators were hired to categorize each image as weird or normal, relying on a majority vote for the final determination. Impressively, human agreement reached 92\%, indicating that, despite subjectivity, there is a clear consensus on what constitutes weirdness within the specific context of the WHOOPS! benchmark.

\subsection{Evaluation}
For our method evaluation on WHOOPS! we use cross-validation with a K-fold setup. Each fold was randomly shuffled with a fixed random seed. In our experiments, we set the number of folds to 3. The training part of the fold is used to tune $w_{ent}$, $w_{con}$ (with fixed $w_{neu}=0$). The result accuracy metric is the average accuracy in all test parts. The weights for the NLI score aggregation $w_{ent}$ and $w_{con}$ were selected on the training part of each fold in the cross-validation pipeline.

\section{Results and Analysis}
The comparison of the described aggregation approaches and NLI models is shown in the Table~\ref{tab:acc}. The \textsf{\small{clust}} method stands out as one of the best performing. This implies that the aggregation of all contradiction scores is crucial, rather than focusing only on extreme values.
In addition, the largest NLI model (\textsf{\small nli-deberta-v3-large}) outperforms all others for all aggregation methods, suggesting that it captures the essence of the problem more effectively.

In each part of the cross-validation fold, the best performing weights were the same ($w_{ent}=1.75$, $w_{con}=-2.0$). Which means that the contradiction score should dominate the entailment score.

Next, we compare our top-performing approach with the baselines in Table~\ref{tab:performance}.  Our method outperforms all other zero-shot techniques and is only marginally behind the fine-tuned approach (\textsf{\small{BLIP2 FlanT5-XXL}}). Surprisingly, both \textsf{\small{InstructBLIP}} baselines outperformed the corresponding \textsf{\small{LLaVA}} models with the same prompt.

\begin{table}[]
\centering
\caption{The performance of different approaches on the WHOOPS! benchmark. Fine-tuned (ft) methods are displayed at the top, while zero-shot (zs) methods are presented at the bottom. Size denotes the number of model parameters. Accuracy as the evaluation metric.}
\begin{tabular}{@{}lccc@{}}
\toprule
\textbf{Model}  & \textbf{\#} & \textbf{Mode} & \textbf{Acc} $\uparrow$ \\ \midrule
BLIP2 FlanT5-XL~\cite{whoops} & 3.94B &ft    & \underline{60.00}              \\
BLIP2 FlanT5-XXL~\cite{whoops} & 12.4B &ft    & \textbf{73.00}              \\ \midrule
BLIP2 FlanT5-XXL~\cite{whoops} & 12.4B &zs     & 50.00\\      
LLaVA 1.6 Mistral 7B~\cite{llava} & 7.57B &zs     & 52.45\\ 
LLaVA 1.6 Vicuna 13B~\cite{llava} & 13.4B &zs     & 56.37\\  
InstructBLIP~\cite{instructblip} & 7B  & zs     &     61.27\\ 
InstructBLIP~\cite{instructblip} & 13B  & zs     &     \underline{62.25}          \\
Ours (nli w/ clust agg) & 7.9B &zs     & \textbf{72.55} \\
\bottomrule
\end{tabular}
\label{tab:performance}
\end{table}

Analysis of predictions regarding the presence of hallucinations in the generated facts and the pair of images with corresponding NLI matrices can be found in Appendicies~\ref{app:hall},~\ref{app:images}.

We acknowledge that GPT-4o, with its impressive accuracy of 79.90\%, could be potentially more effective solution for a considered task. However, GPT-4o is a proprietary system that is much larger than the models employed in our solution. Our goal was to demonstrate how one can utilize a combination of open-source yet efficient visual commonsense models to distinguish odd images. Additionally, to the best of our knowledge, this is the first solution that demonstrates how hallucination can be exploited to solve other problems.

\section{Conclusion}

In this work, we propose a straightforward yet effective approach to visual common sense recognition. Our method exploits an imperfection in LVLMs, causing them to generate hallucinations when presented with unrealistic or strange images. The three-step process involves generating atomic facts, constructing an entailment matrix between these facts, and calculating a common-sense score. This method surpasses all other open-source zero-shot approaches in performance.

\bibliography{sample-ceur}

\newpage
\appendix

\section{Fact Generation Parameters}
\label{app:fact}

In our proposed method \textsf{\small{num\_beams}}, \textsf{\small{num\_beam\_groups}} are set to $5$, and \textsf{\small{diversity\_penalty}} is set to $1.0$. Regarding the penalty, we have found this value to be optimal for adding diversity and preserving the model's ability to follow instructions. The higher values, for example $2.0$ or $5.0$, produce incorrect output according to the instructions. For example, the model produces sentences that are too complex, or ignores the one-sentence constraint, etc.

\section{Cross-Comparison of NLI Models and Aggregation Methods}

\label{app:nli}

\begin{table}[h!]
\begin{center}
\small
\caption{A comparison of various NLI models for distinct aggregation techniques for subset with 5 facts is provided. Accuracy as the evaluation metric.}
\begin{tabular}{@{}lclll@{}}
\toprule
\multirow{2}{*}{\textbf{Model}} & \multirow{2}{*}{\textbf{\#}} & \multicolumn{3}{c}{\textbf{Method}}           \\ \cmidrule(l){3-5} 
                                &                              & \textsf{\small min}            & \textsf{\small absmax}      & \textsf{\small clust}          \\ \midrule
nli-deberta-v3-large            & 304M                         & {\underline{63.73}}   & 62.75 & \textbf{72.55} \\
nli-deberta-v3-base             & 86M                          & {\underline{60.78}} & 55.39 & \textbf{61.76} \\
nli-deberta-v3-small            & 47M                          & \textbf{61.27} &   58.33     & {\underline{60.78}}    \\ \bottomrule
\end{tabular}
\label{tab:acc}
\end{center}
\end{table}

\section{Clustering Algorithm}

\label{app:clust}
\begin{algorithm}[h!]
\begin{center}
\caption{NLI Interval Clustering Aggregation}
\begin{flushleft}
\textbf{Input:} $S_\text{nli}$ – List of weighted and aggregated NLI scores\\
\textbf{Output:} centroids – The minimum centroid NLI score 
\end{flushleft}
\begin{algorithmic}[1]
\STATE kmeans $\leftarrow$ KMeans(clusters=2)
\STATE centroids $\leftarrow$ kmeans.fit($S_\text{nli}$)
\RETURN $\min$(centroids)
\end{algorithmic}
\end{center}
\end{algorithm}

\section{Predictions Analysis}
\label{app:hall}

We analyze how the presence of certain phenomena affects the prediction of our approach. \textsf{\small{digital}} indicates the presence of marker words in the generated facts: ``digital'', ``artistic'', ``rendering''. \textsf{\small{hallucination}} indicates the presence of hallucinations of an object/feature/relationship, manually determined in the generated facts. The relation between the model's decision and the existence of hallucinations or marker words are presented in Table~\ref{tab:digital}.
Our analysis reveals that hallucinations serve as a significant indicator for our model to categorize images as weird. Interestingly, we also discovered that, in certain instances, the LVLM introduces distinctive marker words specifically to denote unusual images.

\begin{table}[ht!]
\centering
\caption{The conditional probability of model prediction being weird given the occurrence of the marker from the corresponding set of marker words are displayed at the top. $\mathbb{P}$(weird $|$ hallucination) = 0.81 indicates the probability that the model predicts weird image given the existence of hallucination in generated atomic facts.}
\begin{tabular}{@{}lc@{}}
\toprule
\textbf{Measure} & \textbf{Value} \\
\midrule
 $\mathbb{P}$(weird $|$ digital) & 0.76 \\
 $\mathbb{P}$(weird $|$ hallucination) & 0.81 \\
 $\mathbb{P}$(weird $|$ hallucination \& digital) & \textbf{0.93} \\
\bottomrule
\end{tabular}
    \label{tab:digital}
\end{table}

\newpage

\section{Pairwise NLI Scores of Atomic Facts with Corresponding Images}
\label{app:images}

\begin{figure*}[h!]
    \centering
    \subfigure[a snow plow driving down a snowy street]
    {\includegraphics[width=1\linewidth]{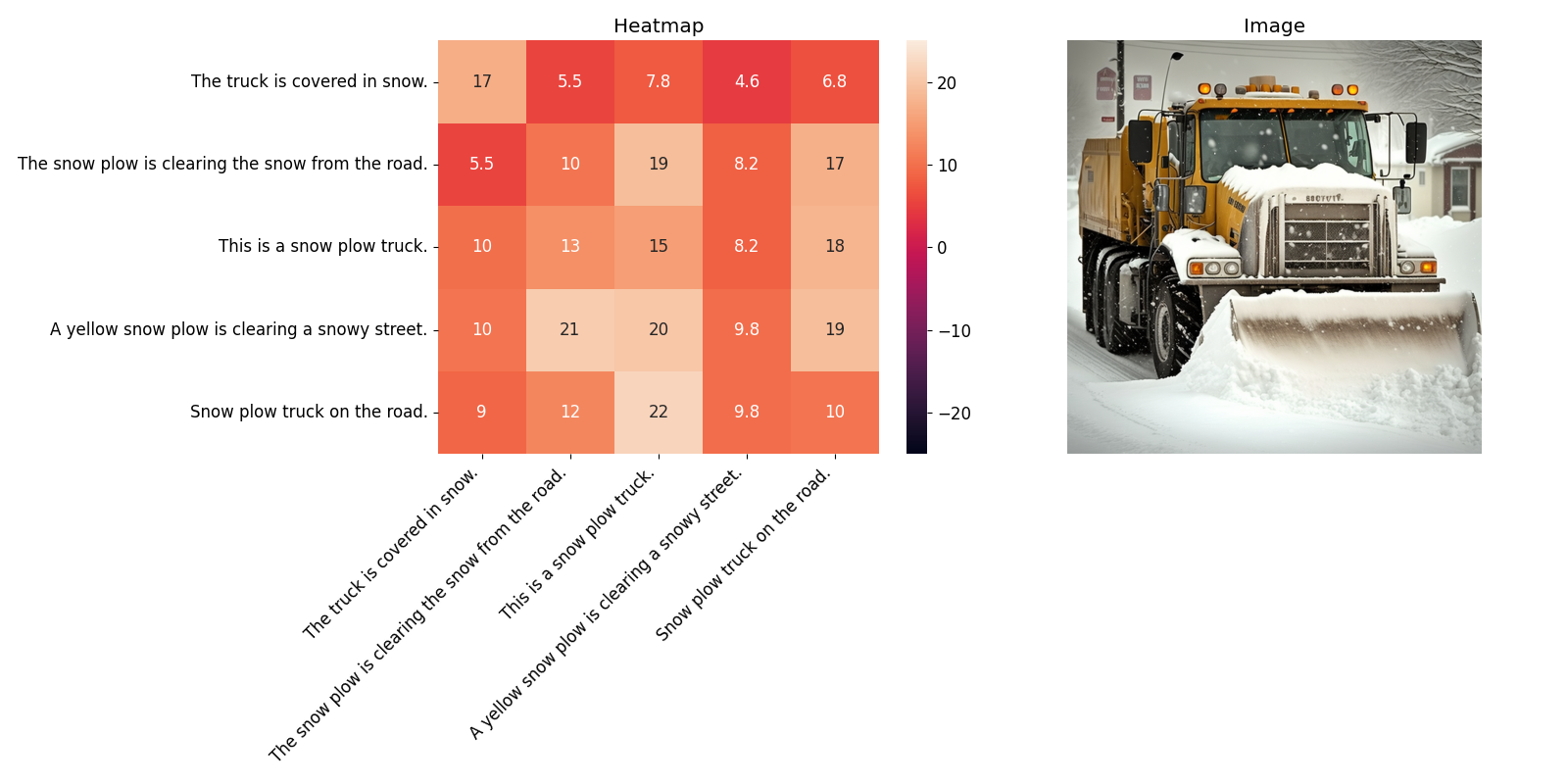}
    }
    \subfigure[a large yellow truck driving through the sand]
    {\includegraphics[width=1\linewidth]{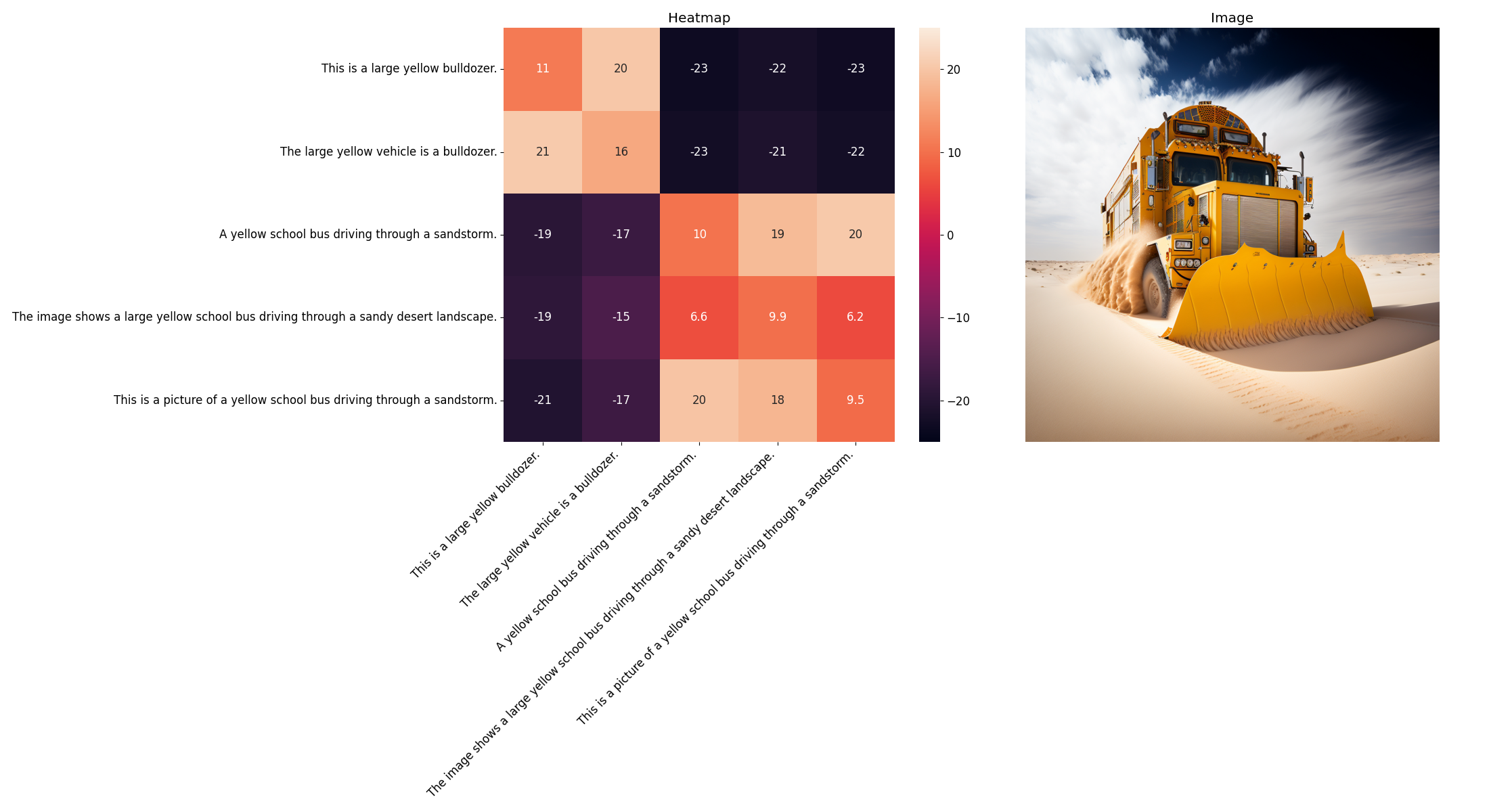}}
\end{figure*} 

\begin{figure*}[h!]
    \centering
    \subfigure[a man sleeping in a bed with a pillow]
    {\includegraphics[width=1\linewidth]{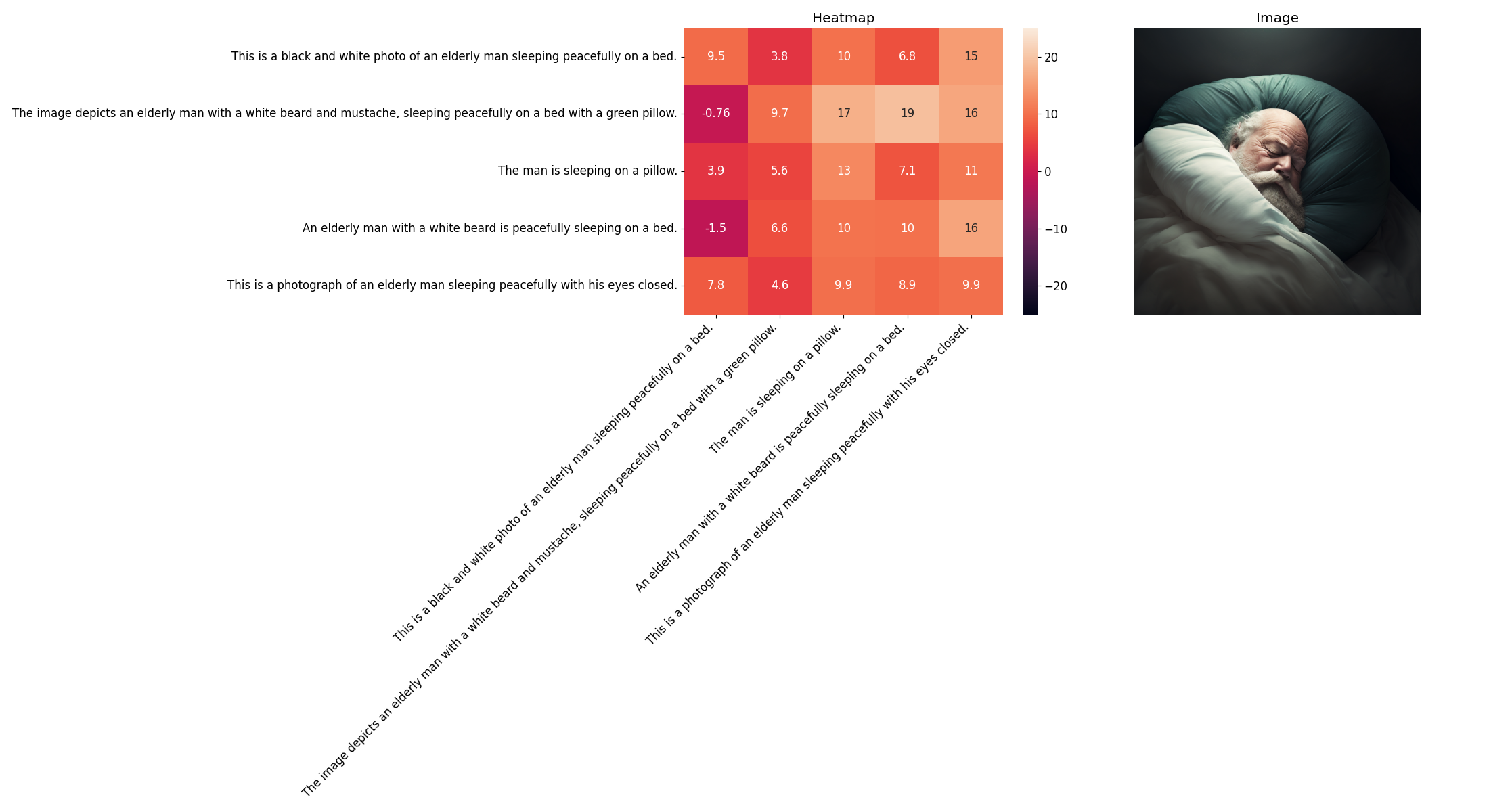}} 
    \subfigure[a man sleeping on a rock]
    {\includegraphics[width=1\linewidth]{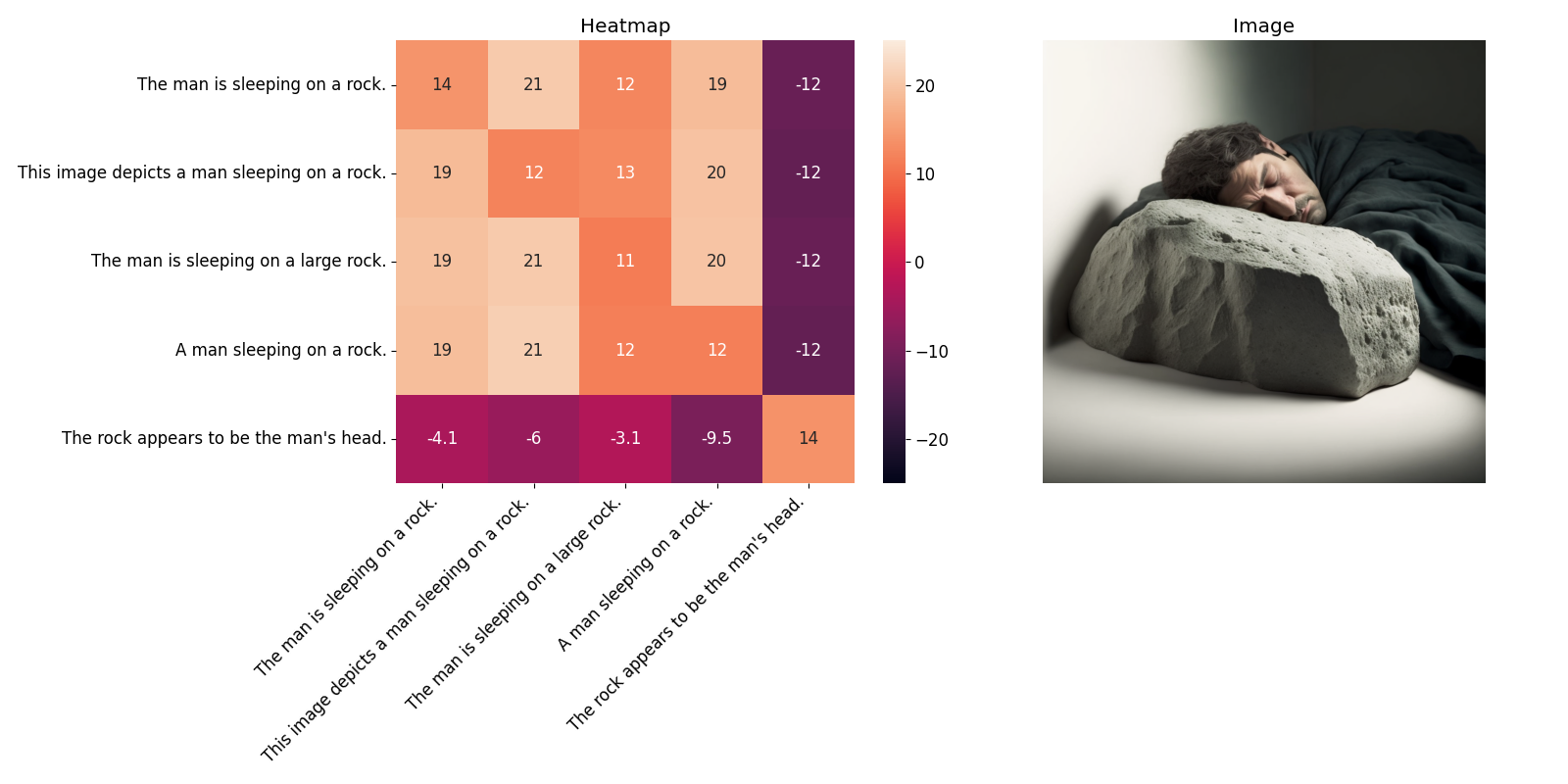}}
\end{figure*} 

\end{document}